\documentclass[10pt,twocolumn,letterpaper]{article}

\usepackage{wacv}
\usepackage{times}
\usepackage{epsfig}
\usepackage{graphicx}
\usepackage{amsmath}
\usepackage{amssymb}
\usepackage{comment}
\usepackage{threeparttable}
\usepackage{multirow}
\usepackage{diagbox}
\usepackage{enumitem}

\newcommand{\rev}[1]{{\color{black} #1}}

\def\wacvPaperID{1278} 

\wacvfinalcopy 

\ifwacvfinal
\usepackage[breaklinks=true,bookmarks=false]{hyperref}
\else
\usepackage[pagebackref=true,breaklinks=true,colorlinks,bookmarks=false]{hyperref}
\fi

\begin{document}
\graphicspath{{figures/}}

\title{Self-Attentive Pooling for Efficient Deep Learning}
\author{Fang Chen$^{1, }$\thanks{Equally contributing authors} , Gourav Datta$^{1, }$\footnotemark[1] , Souvik Kundu$^{2}$, Peter A. Beerel$^{1}$\\
$^{1}$Universiy of Southern California, Los Angeles, USA \ \ \ \  $^{2}$Intel Labs, USA\\
\tt\small\{fchen905, gdatta, pabeerel\}@usc.edu \ \ \ \ souvikk.kundu@intel.com
}

\maketitle

\begin{abstract}
 Efficient custom pooling techniques that can aggressively trim the dimensions of a feature map for resource-constrained computer vision applications have recently gained significant traction.
 However, prior pooling works extract only the local context of the activation maps, limiting their effectiveness. 
 In contrast, we propose a novel non-local \textit{self-attentive pooling} method that can be used as a drop-in replacement to the standard pooling layers, such as max/average pooling or strided convolution. The proposed self-attention module uses patch embedding, multi-head self-attention, and spatial-channel restoration, followed by sigmoid activation and exponential soft-max.
 This self-attention mechanism efficiently aggregates dependencies between non-local activation patches \rev{during down-sampling}.
 Extensive experiments on standard object classification and detection tasks with various convolutional neural network (CNN) architectures demonstrate the superiority of our proposed mechanism over the state-of-the-art (SOTA) pooling techniques. \rev{In particular, we surpass the test accuracy of existing pooling techniques on different variants of MobileNet-V2 on ImageNet by an average of ${\sim}1.2\%$.} 
 \rev{With the aggressive down-sampling of the activation maps in the initial layers (providing up to 22x reduction in memory consumption), our approach achieves $1.43\%$ higher test accuracy compared to SOTA techniques with iso-memory footprints.}
 This enables the deployment of our models in memory-constrained devices, such as micro-controllers \rev{(without losing significant accuracy)}, because the initial activation maps consume a significant amount of on-chip memory for high-resolution images required for complex vision tasks. 
 Our pooling method also leverages channel pruning to further reduce memory footprints. 
 \vspace{-3mm}
 
\end{abstract}


\section{Introduction}
 
In the recent past, CNN architectures have shown impressive stride in a wide range of complex vision tasks, such as object classification \cite{he2016deep} and semantic segmentation \cite{he2018mask}. With the ever-increasing resolution of images captured by modern camera sensors, large activation maps in the initial CNN layers are consuming a large amount of on-chip memory, hindering the deployment of the CNN models on resource-constrained edge devices \cite{datta2022scireports}. Moreover, these large activation maps increase the inference latency, which impedes real-time use cases \cite{datta2022scireports}. Pooling is one of the most popular techniques that can reduce the resolution of these activation maps and aggregate effective features. Historically, pooling layers (either as strided convolution layers or standalone average/max pooling layers) have been used in almost all the SOTA CNN backbones to reduce the spatial size of the activation maps, and thereby decrease the memory footprint of models \cite{inception,densenet,he2016deep}.

Existing pooling techniques aggregate features mainly from the locality perspective.
For example, LIP~\cite{gao2019lip} utilizes a convolution layer to extract locally-important aggregated features.
For relatively simple objects with less diverse feature distribution, it might be sufficient to express them by aggregating local information. But for more complex objects, downsampling feature maps with only local information might be difficult because different local regions of an object might be correlated with each other. For example, an animal's legs can be in different local regions of an image, which might provide useful information to classify the animal as a biped or a tetrapod. Additionally, the features extracted from an object and its background might also be related. For example, with the sea in background, it is highly unlikely that we can find the class `car' in the foreground, and more likely that we can find classes like `boat' or `ship' in the foreground. 
 
To reduce the significant on-chip memory consumed by the initial activation maps, large kernel sizes and strides are often required in the pooling layers to fit the models in resource-constrained devices. This might lead to loss of feature information when only leveraging locality for aggregation. Moreover, recently proposed in-sensor \cite{kodukula2020sensors,chen2020pns} and in-pixel \cite{datta2022scireports,datta2022hsipip,dattap2mdetrack} computing approaches can benefit from aggressive bandwidth reduction in the initial CNN layers via down sampling. We hypothesize that the accuracy loss typically associated with aggressive down-sampling can be minimized by considering both local and non-local information during down-sampling.

To explore this hypothesis, we divide the activation map into patches and propose a novel non-local self-attentive pooling method to aggregate features and capture long-range dependencies across different patches.
The proposed method consists of a patch embedding layer, 
a multi-head self-attention layer, a spatial-channel restoration layer, followed by a sigmoid and an exponential activation function.  The patch embedding layer encodes 
each patch into a one-pixel token that consists 
of multiple channels.
The multi-head self-attention layer models the long-range dependencies between different patch tokens. The spatial-channel restoration layer helps in decoding and restoring the patch tokens to non-local self-attention maps. The sigmoid and exponential activation functions rectify and amplify the non-local self-attention maps, respectively. Finally, the pooled activation maps compute  the patch-wise average of the element-wise multiplication of the input activation maps and the non-local self-attention maps. 

Our method surpasses the test accuracy (mAP) of all existing pooling techniques in CNNs for a wide range of on-device object recogniton (detection) tasks, particularly when the initial activation maps need to be significantly down-sampled for memory-efficiency. Our method can also be coupled with structured model compression techniques, such as channel pruning, that can further reduce the compute and memory footprint of our models.



In summary, the key highlights of this paper can be summarized as 
\begin{itemize}[itemsep=0pt,parsep=0pt,topsep=0pt,partopsep=0pt]
\item Inspired by the potential benefits of non-local feature aggregation, we propose the use of multi-head self-attention to aggressively downsample the activation maps in the initial CNN layers that consume a significant amount of on-chip memory. 
\item We propose the use of spatial channel restoration, weighted averaging, and custom activation functions in our self-attentive pooling approach. Additionally, we jointly optimize our approach with channel pruning to further reduce the memory and compute footprint of our models. 
 \item We demonstrate the memory-compute-accuracy (mAP) trade-off benefits of our proposed approach through extensive experiments with different on-device CNN architectures on both object recogniton and detection tasks, and comparisons with existing pooling and memory-reduction approaches. Moreover, we provide visualization maps obtained by our non-local pooling technique which provides deeper insights on the efficacy of our approach.
\end{itemize}
 

\begin{figure*}[htbp]
\centerline{\includegraphics[scale=0.6]{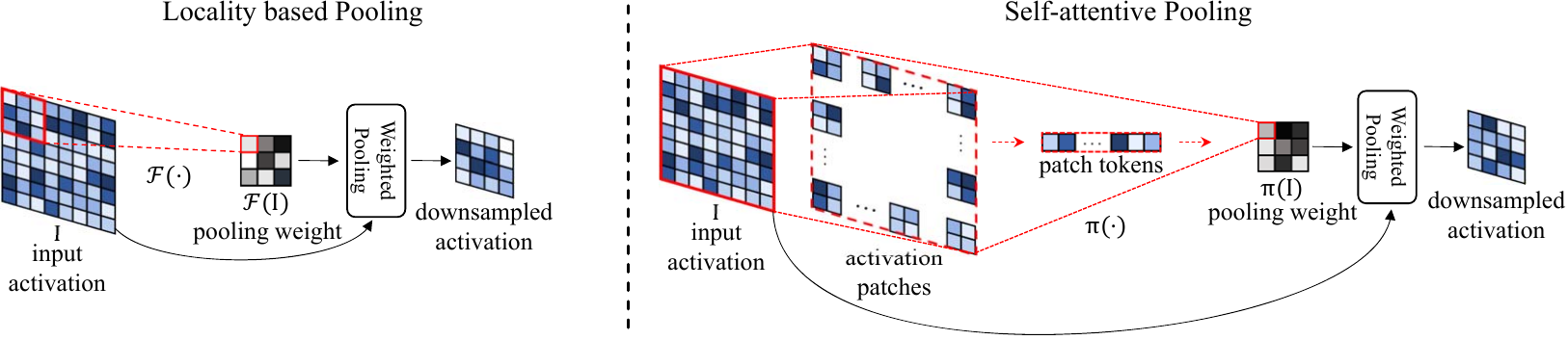}}
\caption{
Illustration of locality based pooling and non-local self-attentive pooling.
The pooling weight has the same shape with the input activation $I$, of which only a local region is displayed in this figure. $\mathcal{F}(\cdot)$ denotes the locality based pooling and $\pi(\cdot)$ denotes the proposed non-local self-attentive pooling. For the locality based pooling, each pooling weight has limited sensitive field as shown in the red box. For the proposed non-local self-attentive pooling, the input activation is divided to several patches and encoded into a series of patch tokens. Based on these patch tokens, the pooling weights have global view, which makes it superior for capturing long-range dependencies and aggregating features.}
\label{fig:pooling illsutration}
\end{figure*}

\section{Related Work}\label{sec:related_work}


\subsection{Pooling Techniques}

Most popular CNN backbones consist of pooling layers for feature aggregation. For example, VGG~\cite{simonyan2014very}, Inception \cite{inception} and DenseNet \cite{densenet} use either average/max pooling layers, while ResNet \cite{he2016deep}, MobileNet \cite{sandler2018mobilenetv2}, and their variants use convolutions with stride greater than 1 as trainable pooling layers in a hierarchical fashion at regular locations for feature down-sampling. 
However, these naive pooling techniques might not be able to extract useful and relevant features, particularly when the pooling stride needs to be large for significant down-sampling. This has resulted in a plethora of novel pooling layers that have been introduced in the recent past. 

In particular, mixed \cite{mixed_pool} and hybrid pooling \cite{hybrid_pooling} use a combination of average and max pooling, which can be learnt during training. L$_p$ pooling \cite{norm_pooling} extracts features from the local window using the L$_p$ norm, where the parameter \textit{p} can be learnt during training. Another work proposed detail-preserving pooling (DPP) \cite{dpp} where the authors argued that there are fine-grained details in an activation map that should be preserved while the redundant features can be discarded. However, the detail score is an arbitrary function of the statistics of the pixel values in the receptive field that may not be optimal. 
A more recent work introduced local importance pooling (LIP) \cite{gao2019lip} that uses a trainable convolution filter that captures the local importance of different receptive fields and is used to scale the activations 
before pooling. Gaussian based pooling~\cite{kobayashi2019gaussian} formulates the pooling operator as a probabilistic model to flexibly represent the activation maps. RNNPool~\cite{NEURIPS2020_ebd9629f} uses recurrent neural networks (RNNs) to aggregate features of large 1D receptive fields across different dimensions. \rev{In order to extract context-aware rich features for fine-grained visual recognition, another recent work \cite{behera2021context} called CAP proposed a novel attentive pooling that correlates between different regions of the convolutional feature map to help discriminate between subcategories and improve accuracy. In particular, CAP is applied late in the network (after all convolutional layers) and is not intended to reduce the model's memory footprint, in contrast to our work which applies pooling to down-sample the large activation maps early in the network. 
Interestingly, CAP transforms each feature using a novel form of attention (involving only query and key) rather than the traditional self-attention module adopted in this work.
%
%
Lastly, 
while CAP uses bi-linear pooling, global average pooling, and LSTM, our approach uses a patch embedding, spatial channel-restoration, and weighted pooling.} 



\subsection{Model Compression}

Pruning is one of the well-known forms of model compression \cite{kundu2022bmpq, han2015deep, hinton2015distilling, kundu2021analyzing} that can effectively reduce the DNN inference costs \cite{han2015deep, kundu2022towards}.  
A recent surge in pruning methods has opened various methods for pruning subnetworks, including iterative magnitude pruning (IMP) \cite{frankle2018lottery}, reinforcement learning driven methods \cite{he2018amc}, additional optimization based methods \cite{li2019admm}. However, these methods require additional training iterations and thus demand significantly more training compute cost. In this work, we adopt a more recent method of model pruning, namely \textit{sparse learning} \cite{kundu2021dnr}, that can effectively yield a pruned subnetwork while training from scratch. In particular, as this method always updates a sparse subnetwork to non-zero and ensures meeting the target pruning ratio, we can safely avoid the fine tuning stage yet obtain good accuracy. Readers interested in pruning and sparse learning can refer to \cite{hoefler2021sparsity} for more details. Recently, neural architecture search \cite{lin2020mcunet} also enabled significant model compression, particularly for memory-limited devices. A recent work \cite{lin2021mcunetv2} proposed patch-based inference and network redistribution to shift the receptive field to later stages to reduce the memory overhead. 

\subsection{Low-Power Attention-based Models}
 \rev{There are a few self-attention-based transformer models in the literature that aim to reduce the compute/memory footprint for edge deployments.
 MobileVit~\cite{mehta2022mobilevit} proposed a light-weight and general-purpose vision transformer, combining the strengths of CNNs and ViTs.
 LVT~\cite{yang2022lite} proposed two enhanced self-attention mechanisms for low- and high-level features to improve the models performance on mobile devices. MobileFormer \cite{chen2022mobile} parallelized MobileNet and Transformer with a two-way bridge for information sharing, which achieved SOTA performance in accuracy-latecy trade-off on ImageNet.
 For other vision tasks, such as semantic segmentation and cloud point downsampling, recent works have proposed transformer-based models for mobile devices. For example, LighTN~\cite{wang2022lightn} proposed a single-head self-correlation module to aggregate global contextual features and a down sampling loss function to guide training for cloud point recognition.
 TopFormer~\cite{zhang2022topformer} utilized tokens pyramid from various scales as input to generate scale-aware semantic features for semantic segmentation.
}
\section{Background}\label{sec:prelims}
 In this section, we explain the multi-head self-attention~\cite{vaswani2017attention} module that was first introduced by the ViT architecture~\cite{dosovitskiy2020image} in computer vision.
 
 In ViT, the input image $I\in\mathbb{R}^{H \times W \times C}$ is reshaped into a sequence of non-overlapping patches $I_p\in\mathbb{R}^{(\frac{H \cdot W}{P^2})\times(P^2 \cdot C)}$, where $(H\times W)$ is the size of the input RGB image, and $C$ is the number of channels, and $P^2$ is the number of pixels in a patch.
 The flattened 2D image patches are then fed into the multi-head self-attention module. 
 Specifically, the patch sequence $I_p$ are divided into $m$ heads $I_p=\{I_p^1, I_p^2, ..., I_p^m\}\in\mathbb{R}^{N\times\frac{C_p}{m}}$, where $N=(\frac{H \cdot W}{P^2})$ is the number of patches and $C_p=P^2 \cdot C$ is the number of channels in $I_p$.
 These tokens are fed into the multi-head self-attention module $MSA(\cdot)$:
 \begin{small}
 \begin{eqnarray}
 {I_a = LN(MSA(I_p)) + I_p, } \label{eq:MSA}
 \end{eqnarray}
 \end{small}where $LN(\cdot)$ is the layer normalization~\cite{wang2019learning, baevski2018adaptive}.
 
 In the $j$-th head, the token series $I_p^j\in\mathbb{R}^{N\times\frac{C_p}{m}}$ is first projected onto $L(I_p^j)\in\mathbb{R}^{N\times d_k}$ by a linear layer.
 Then three weight matrices $\{W^q, W^k, W^v\} \in \mathbb{R}^{D\times d_k}$ are used to obtain the query, key, and value tokens as $Q^j=W^qL(I_p^j), K^j=W^kL(I_p^j), V^j=W^vL(I_p^j)$, respectively.
 $D$ is the hidden dimension and $d_k=D/m$.
 The output $I_a^j\in\mathbb{R}^{N\times D}$ of the self-attention layer is given by:
 \begin{small}
 \begin{eqnarray}
 {I_a^j = \textit{softmax}(\frac{Q^jK^{j^T}}{\sqrt{d_k}})V^j. } \label{eq:self-attention}
 \end{eqnarray}
 \end{small}Finally, the results of the $m$ heads are concatenated and back projected onto the original space:
 \begin{small}
 \begin{eqnarray}
 {I_a = \textit{concat}(I_a^1, I_a^2, ..., I_a^m)W^O, } \label{eq:x_attn}
 \end{eqnarray}
 \end{small}where $W^O\in\mathbb{R}^{C_p\times D}$ is the projection weight and the final output $I_a\in\mathbb{R}^{N\times C_p}$.

\section{Proposed Method}
\label{sec: proposed method}
The weights of local pooling approaches are associated with only a local region of the input feature maps as shown in Fig.~\ref{fig:pooling illsutration}. 
These pooling approaches are limited by the locality of the convolutional layer, and need a large number of layers to acquire a large sensitive field.  
To mitigate this issue,  we can intuitively encode the global and non-local information into the pooling weights, as shown in Fig.~\ref{fig:pooling illsutration}.
To realize this intuition, we propose a form of self-attentive pooling that is based on a multi-head self-attention mechanism which captures the non-local information as self-attention maps that perform feature down-sampling. 
Then, we jointly optimize the proposed pooling method with channel pruning to further reduce the memory footprint of the entire CNN models.

\begin{figure}[htbp]
\centerline{\includegraphics[scale=0.98]{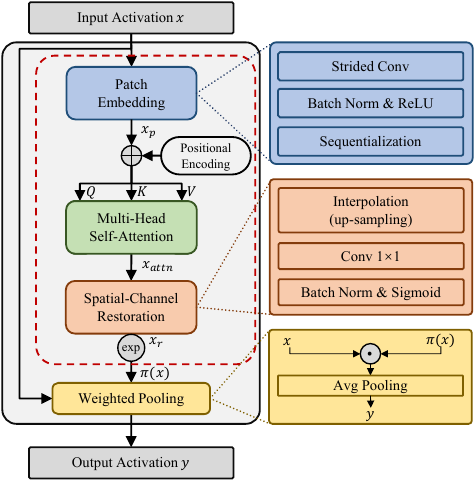}}
\caption{
 Architecture of the non-local self-attentive pooling.
 }
\label{fig:Non-Local-Pooling}
\vspace{-4mm}
\end{figure}

\subsection{Non-Local Self-Attentive Pooling}\label{subsec:nlp}
 The overall structure of the proposed method is shown in Fig.~\ref{fig:Non-Local-Pooling}.
 It consists of four main modules: patch embedding, multi-head self-attention, spatial-channel restoration, and weighted pooling.
 
 1) \textbf{Patch embedding} is used to compress spatial-channel information.
We use a strided convolution layer to encode and compact local information 
for different patches along the spatial and channel dimensions of the input.
More precisely, the input to the embedding is a feature map denoted as $x \in{\mathbb{R}^{h \times w \times c_x}}$  with resolution $(h\times w)$ and $c_x$ input channels. The output of the embedding is a token series $x_p\in{\mathbb{R}^{(\frac{h \cdot w}{\epsilon^2_p}) \times (\epsilon_r \cdot c_x)}}$, where $\epsilon_p$ is the patch size and
 $\epsilon_r$ sets the number of output channels as $\epsilon_r \cdot c_x$.
The patch embedding consists of a strided convolution layer with kernel size and stride both equal to $\epsilon_p$ followed by a batch norm layer and a ReLU function~\cite{agarap2018deep}.

 For each patch indexed by $[n_i,n_j]$, the patch embedding layer output can be formulated as:
 \begin{small}
 \begin{eqnarray}
 x_p[n_i,n_j]=
 \phi_{relu}\left(\sum_{i=0}^{\epsilon_p}\sum_{j=0}^{\epsilon_p}{w^c_{i,j} \cdot x_{(n_i\cdot\epsilon_p+i, n_j\cdot\epsilon_p+j)}} + b^c\right)  \label{eq:patch embedding}
 \end{eqnarray}
\end{small}

\noindent
where $w^c, b^c$ are the weight and bias of the convolution kernel, respectively, and $\phi_{relu}$ denotes the ReLU activation function.
After patch embedding, a learnable positional encoding~\cite{dosovitskiy2020image} is added to the token series $x_p$ to mitigate the loss of positional information caused by sequentialization.
 
 2) \textbf{Multi-head self-attention} is used to model the long-range dependencies between different patch tokens.
 While the input patch token series $x_p$ is fed into the module, the output $x_{attn}$ is a self-attentive token sequence with the same shape as $x_p$.
 
 3) \textbf{Spatial-channel restoration} decodes spatial and channel information from the self-attentive token sequence $x_{attn}$.
 The token sequence $x_{attn}\in{\mathbb{R}^{(\frac{h \cdot w}{\epsilon^2_p}) \times (\epsilon_r \cdot c_x)}}$ is first reshaped to $\mathbb{R}^{\frac{h}{\epsilon_p}\times\frac{w}{\epsilon_p}\times(\epsilon_r\cdot C)}$, and then expanded to the original spatial resolution $(h, w)$ via bilinear interpolation. A subsequent convolutional layer with $1 \times 1$ kernel size  projects the output to the same number of channels $c_x$ as the input tensor $x$.
 A batch norm layer normalizes the response of the output attention map $x_r\in\mathbb{R}^{h \times w \times c_x}$.
 A sigmoid function is then used to rectify the output range of $x_r$ to [$0{,}1$], followed by an exponential function to amplify the self-attentive response.

4) \textbf{Weighted pooling} is used to generate the down-sampled output feature map from the output of the spatial-channel restoration block, denoted as $\pi(x)$ in Fig. \ref{fig:Non-Local-Pooling}. In particular, assuming a kernel and stride size of ($s{\times}s$) in our pooling method, and considering a local region in $x$ from $(p, q)$ to $(p+s, q+s)$, the pooled output corresponding to this region can be estimated as
 \begin{small}
 \begin{eqnarray}\label{eq:local_pool}
O = \frac{\sum_{i=p}^{i=p+s} \sum_{j=q}^{j=q+s}\pi_{i,j}(x)x_{i,j}}{\sum_{i=p}^{i=p+s} \sum_{j=q}^{j=q+s}\pi_{i,j}(x)}
 \end{eqnarray}
 \end{small}
where $\pi_{i,j}(x)$ denotes the value of $\pi(x)$ at the index (i,j). Similarly, the whole output activation map can be estimated from each local region separated with a stride of $s$. 


\subsection{Optimizing with Channel Pruning}

To further reduce the activation map dimension we leverage the popular channel pruning \cite{kundu2021dnr} method. In particular, channel pruning ensures all the values in some of the convolutional filters to be zero. This in turn makes the associated activation map channels redundant. 
Let us assume a layer $l$ with corresponding 4D weight tensor $\pmb{{\theta}}_l \in \mathbb{R}^{M \times N \times h \times w}$. Here, $h$ and $w$ are the height and width of 2D kernel of the tensor, with $M$ and $N$ representing the number of filters and channels per filter, respectively. To perform channel pruning of the layer weights, we first convert the weight tensor $\pmb{{\theta}}_l$ to a 2D weight matrix, with $M$ and $N \times h \times w$ being the number of rows and columns, respectively. We then partition this matrix into $N$ sub-matrices of $M$ rows and $h \times w$ columns, one for each channel. To rank the importance of the channels, for a channel $c$, we then compute the Frobenius norm (F-norm) of its associated sub-matrix, meaning effectively compute $O_l^c$ = $\lvert{\pmb{{\theta}}_l^{:,c,:,:}}\rvert^{2}_F$. 
Based on the fraction of non-zero weights that need to be rewired during an epoch $i$, denoted by the pruning rate $p_i$, we compute the number of channels that must be pruned from each layer, ${c}^{p_i}_l$, and prune the ${c}^{p_i}_l$ channels with the lowest F-norms. 
We then leverage the normalized momentum contributed by a layer's non-zero channels to compute its layer importance that are then used to measure the number of zero-F-norm channels $r_l^i \geq 0$ that should be re-grown for each layer $l$. 
Note that we first pre-train CNN models with our self-attentive pooling, and then jointly fine-tune our pooled models with this channel pruning technique. While the pooling layers are applied to all down-sampling layers, the channel pruning is only applied on the initial activation maps (only in the first stage in CNN backbones illustrated in Fig. 3) to maximize its' impact of reducing the memory footprint of the models.
\vspace{-6mm}
\section{Self-Attentive Pooling in CNN Backbones}
\label{sec: using pooling in backbone networks}

\begin{figure}
\centerline{\includegraphics[scale=0.48]{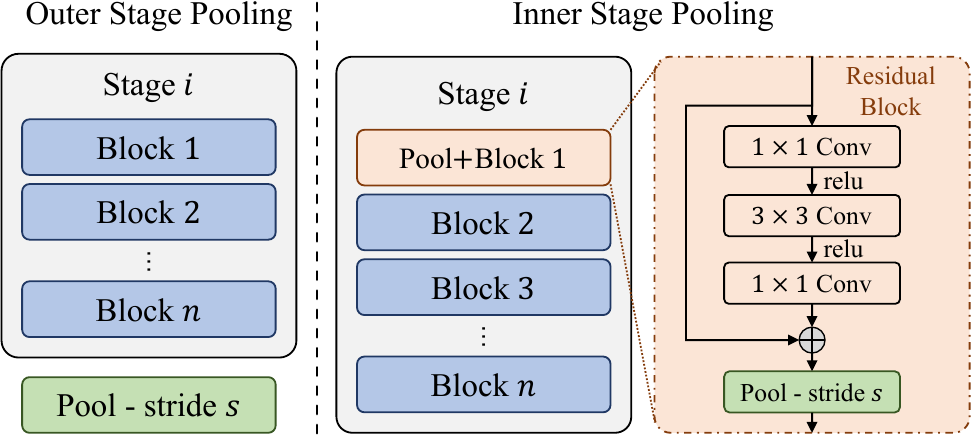}}
\caption{
 Illustration for two ways of using pooling methods.
 }
\vspace{-3mm}
\label{fig:pooling2ways}
\end{figure}

 The proposed pooling method can be used in any backbone networks, such as VGG~\cite{simonyan2014very}, MobileNet~\cite{sandler2018mobilenetv2} and ResNet~\cite{he2016deep}.
 Generally, a backbone network can be roughly divided into several stages and, the down-sampling layer (either as a strided convolution or max/average pooling), if present in a stage, is only applied at the first block. 
 Specifically, there are two ways to replace this down-sampling layer with our (or any other SOTA) pooling method in the backbone network, i.e., outer-stage pooling and inner-stage pooling, as shown in Fig.~\ref{fig:pooling2ways}.
 Outer stage pooling means that the activation is down-sampled by the pooling layer after each stage, which helps to reduce the size of the final output activation map in each stage and customize the pooling layer to learn the stage information.
 Inner stage pooling means the activation is down-sampled after the first block of each stage, which helps to reduce the initial activation map.
 We optimize the use of these pooling methods for each backbone evaluated, as specified in Section 6.1.


\section{Experiments}
\label{sec:experiments}

 
\subsection{Experimental Setup}

The proposed pooling method is compared to several pooling methods, such as strided convolution, LIP, GaussianPool, and RNNPool. All these methods are widely used in deep learning and, to the best of our knowledge, yield SOTA performance.
Our proposed method is implemented in PyTorch  with the hyper-parameter settings, along with for those we compare with, listed in Table~\ref{tab:hyper parameter settings}. Specifically, we evaluate the pooling approaches on two compute- and memory-efficient backbone networks. MobileNetv2 and ResNet18. For both, we keep the same pooling settings except the first pooling layer, where we employ aggressive striding for memory reduction. For example, in MobileNetV2, we use strides $(s1, 2, 2, 2, 1, 2, 1)$, where $s1\in\{1, 2, 4\}$. More details are in supplementary materials. 

\begin{table}
\caption{Hyper parameter settings of different pooling techniques}
\label{tab:hyper parameter settings}
\begin{center}
\setlength{\tabcolsep}{2mm}{
\begin{tabular}{lc}
\hline
\hline
Methods & Parameter Settings \\
\hline

Strided Conv. & kernel size: $3\times3$ \\

LIP & kernel size: $1\times1$ \\


\hline
\multirow{2}{*}{Ours} & $\epsilon_p\in\{1, 2, 4, 8\},$ \\
& $\epsilon_r\in\{0.25, 1\}, m: 2$ \\
Channel Pruning &  $2\times$ \\

\hline
\hline
\vspace{-2mm}
\end{tabular}
}
\end{center}
\vspace{-6mm}
\end{table}

 To evaluate the performance of the pooling methods on multi-object feature aggregation, we use two object detection frameworks, namely SSD \cite{liu2016ssd} and Faster R-CNN \cite{ren2015faster}. To holistically evaluate the pooling methods, we use three image recognition datasets, namely STL-10, VWW, and ImageNet, which have varying complexities and use-cases. Their details are in supplementary materials. To evaluate on the multi-object detection task, we use the popular Microsoft COCO dataset~\cite{coco}. 
Specifically, we use an image resolution of $300\times300$ for the SSD framework, the same as used in~\cite{liu2016ssd}, $608\times608$ for the YoloV3 framework, the same as used in~\cite{redmon2018yolov3}, and $1333\times800$ for the Faster RCNN framework.
 Eight classes related to autonomous driving which includes \{'person', 'bicycle', 'car', 'motorcycle', 'bus', 'train', 'truck', 'traffic light'\} are used for our experiments.
 We evaluate the performance of each pooling method using mAP averaged for IoU $\in\{0.5, 0.75, [0.5:0.05:0.95]\}$, denoted as mAP@0.5, mAP@0.75 and mAP@[0.5, 0.95], respectively. We also report the individual mAPs for small (area less than $32^2$ pixels), medium (area between $32^2$ and $96^2$ pixels), and large (area more than $96^2$ pixels) objects. 


\begin{table}
\caption{Comparison of different pooling methods for different CNN backbones on STL10 dataset.}
\vspace{-2mm}
\label{tab: comparison on stl10}
\begin{center}
\setlength{\tabcolsep}{0.5mm}{
\begin{threeparttable}
\begin{tabular}{l|c|c|c}
\hline
\hline
Metrics & \multicolumn{3}{c}{Top 1 Acc. (\%)} \\
\hline

\diagbox[]{Methods$^{*}$}{$1^{st}$ Pool Stride} 
  & $s1=1$ & $s1=2$ & $s1=4$\\
\hline

Strided Conv.--MobileNetV2
    & 79.69 & 72.49 & 36.49 \\
LIP--MobileNetV2
    & 79.16 & 68.23 & 36.50 \\
GaussianPool--MobileNetV2
     & 81.50 & 74.56 & 33.31 \\
RNNPool--MobileNetV2 
     & 81.62   & 74.62 & 37.42  \\
  Ours--MobileNetV2
    & 81.75 & \textbf{75.39} & \textbf{40.66} \\
Ours+CP$^{**}$--MobileNetV2
    & \textbf{82.38} & 74.12 & 37.44 \\
\hline
Strided Conv.--MobileNetV2-0.35x
    & 69.89 & 63.72 & 31.45 \\
    LIP--MobileNetV2-0.35x
    & 73.02 & 65.91 & 33.97 \\
GaussianPool--MobileNetV2-0.35x
     & 71.67 & 67.88 & 35.03 \\
 RNNPool--MobileNetV2-0.35x & 72.90 & 67.41 & 35.09 \\
Ours--MobileNetV2-0.35x
    & \textbf{77.99} & \textbf{69.30} & \textbf{36.68} \\
Ours+CP-MbNetV2-0.35x & 77.43 & 68.08 & 33.30 \\
\hline
Strided Conv.--ResNet18
    & 79.80 & 76.05 & 66.49 \\
LIP--ResNet18
    & 81.94 & 80.53 & 78.55 \\
 GaussianPool--ResNet18
     & 81.57 & 78.70 & 74.61 \\
RNNPool--ResNet18  
     & 81.80     & 80.26     & 78.62    \\
Ours--ResNet18
    & 82.25 & \textbf{81.11} & \textbf{79.39} \\
Ours+CP--ResNet18
    & \textbf{82.68} & 79.81 & 76.19 \\
\hline
\hline
\end{tabular}
\begin{tablenotes}
    \footnotesize{
    \item * Methods are named by pooling method's name-backbone's name. MbNetV2 indicates the MobileNetV2 backbone network. 'Ours' indicates the standard proposed pooling method.
    \item ** 'Ours+CP' is the proposed method with 2$\times$ channel pruning in the backbone network before the $1^{st}$ pooling layer. 
    }
\end{tablenotes}
\end{threeparttable}
}
\vspace{-6mm}
\end{center}
\end{table}

\begin{table}
\caption{Comparison of different pooling methods for MobileNetV2-0.35X on VWW dataset.}
\label{tab: comparison on vww}
\begin{center}
\setlength{\tabcolsep}{0.3mm}{
\begin{threeparttable}
\begin{tabular}{l|c|c|c}
\hline
\hline
Metrics & \multicolumn{3}{c}{Top 1 Acc. (\%)} \\
\hline

\diagbox[]{Methods$^{*}$}{$1^{st}$ Pool Stride} 
  & $s1=1$ & $s1=2$ & $s1=4$\\
\hline

Strided Conv.--MobileNetV2-0.35x
    & 91.72 & 83.52 & 78.83 \\
LIP--MobileNetV2-0.35x
    & 91.24 & 83.30 & 79.48 \\
GaussianPool--MobileNetV2-0.35x
    & 91.09 & 82.81 & 79.51 \\
RNNPool--MobileNetV2-0.35x 
    & 90.85 & 83.41  & 79.20  \\
\hline
Ours--MobileNetV2-0.35x
    & \textbf{91.86} & \textbf{83.87} & \textbf{80.21}\\
Ours+CP$^{**}$-MbNetV2-0.35x & 91.60 & 82.46 & 76.11 \\
\hline
\hline
\end{tabular}
\end{threeparttable}
}
\vspace{-7mm}
\end{center}
\end{table}

\begin{table*}[t]
\caption{Comparison on COCO dataset.}
\label{tab: comparison on coco}
\begin{center}
\setlength{\tabcolsep}{1mm}{
\begin{threeparttable}
\begin{tabular}{l|c|c|c|c|c|c|c}
\hline
\hline

\multirow{2}{*}{Framework} & \multirow{2}{*}{Methods} & \multicolumn{6}{c}{mAP} \\
\cline{3-8}
& & @0.5 & @0.75 & @[0.5,0.95] & @large & @medium & @small \\
\hline

\multirow{6}{*}{SSD}
& Strided Conv.--MobileNetV2 
    & 36.30 & 23.00 & 21.90 & 44.60 & 14.40 & 0.80 \\
& LIP--MobileNetV2 
    & 37.50 & 23.10 & 22.30 & 44.80 & 15.30 & \textbf{0.90} \\
& GaussianPool--MobileNetV2 
    & 37.00 & 24.00 & 22.80 & 46.50 & 16.00 & 0.70 \\
& Ours--MobileNetV2 
    & \textbf{38.00} & \textbf{24.50} & \textbf{23.30} & \textbf{47.00} & \textbf{16.50} & 0.80 \\
\cline{2-8}
& Strided Conv.--ResNet18 
    & 38.80 & 24.70 & 23.40 & 47.00 & 15.70 & 1.10 \\
& LIP--ResNet18 
    & 40.60 & 25.10 & 24.20 & 47.80 & 18.00 & \textbf{1.70} \\
& GaussianPool--ResNet18 
    & 40.40 & 24.90 & 24.10 & 47.20 & 17.70 & 1.40 \\

& Ours--ResNet18 
    & \textbf{41.60} & \textbf{25.40} & \textbf{24.90} & \textbf{48.80} & \textbf{19.30} & 1.60 \\




\hline

\multirow{3}{*}{Faster RCNN} 
& Strided Conv.--ResNet18 
    & 63.60 & 40.80 & 38.70 & \textbf{52.70} & 36.70 & 21.00 \\
& LIP--ResNet18 
    & 65.30 & 42.00 & 39.90 & 52.10 & 39.00 & \textbf{23.90} \\
& GaussianPool--ResNet18 
    & 55.30 & 33.10 & 31.80 & 44.40 & 29.10 & 16.00 \\

& Ours--ResNet18 
    & \textbf{65.50} & \textbf{42.60} & \textbf{40.00} & 51.50 & \textbf{39.90} & 22.80 \\

\hline
\hline
\end{tabular}
\end{threeparttable}
}
\vspace{-5mm}
\end{center}
\end{table*}

 \subsection{Accuracy \& mAP Analysis}


 The experimental results on the image recognition benchmarks are illustrated in Table~\ref{tab: comparison on stl10},~\ref{tab: comparison on vww}, and ~\ref{tab: comparison on imagenet}, where each pooling method is applied on the different backbone networks described in Section 6.1.2. Note that the resulting network is names as `pooling method's name'--`backbone network's name'.
 For example, `Strided Conv.--MobileNetV2' means we use the strided convolution as the pooling layer in the MobileNetV2 backbone network.
 
 On the STL10 dataset, when evaluated with the MobileNetV2 and ResNet18 backbone network, the proposed method outperforms the existing pooling approaches by approximately $0.7\%$ for $s1{=}1$.
 In contrast, in ImageNet, the accuracy gain ranges from $0.86\%$ to $1.66\%$ (\rev{$1.2\%$ on average}) for $s1{=}1$. Since VWW is a relatively simple task, the accuracy gain of our proposed method is only $0.14{\sim}0.7\%$ across different values of $s1$. Further analysis of the memory-accuracy trade-off with channel pruning and other $s1$ values is presented in Section \ref{subsec:comp_mem}. 
 
 
 

 The object detection experimental results for $s1{=}1$ are listed in Table~\ref{tab: comparison on coco}.
 When evaluated on SSD framework, our proposed method outperforms the SOTA pooling approach by $0.5\%\sim1\%$ for mAP@0.5, $0.3\%\sim0.5\%$ for mAP@0.75 and $0.5\%\sim0.8\%$ for mAP@[0.5,0.95], which illustrates the superiority of our method on multi-object feature aggregation.
 When evaluated on Faster RCNN framework, the proposed method also achieves the state-of-the-art performance on mAP@0.5, mAP@0.75 and mAP@[0.5, 0.95] with approximately $0.1\%\sim0.6\%$ mAP gain.
 
 All results, except those for ImageNet and COCO (due to compute constraints), are reported as the mean from three runs with distinct seeds, and the variance from these runs is $<$0.1\% which is well below our accuracy gains.
 
 \begin{table}
\caption{Comparison of different pooling methods for MobileNetV2-0.35x on ImageNet dataset.}
\label{tab: comparison on imagenet}
\begin{center}
\setlength{\tabcolsep}{0.5mm}{
\begin{threeparttable}
\begin{tabular}{l|c|c}
\hline
\hline
Metrics & \multicolumn{2}{c}{Top 1 Acc. (\%)} \\
\hline

\diagbox[]{Methods}{$1^{st}$ Pool Stride} 
  & $s1=1$ & $s1=2$ \\
\hline

Strided Conv.--MobileNetV2
    & 70.02 & 60.18  \\
LIP--MobileNetV2
    & 71.62 & 61.86  \\
GaussianPool--MobileNetV2
     & 72.02 & 61.24  \\
RNNPool--MobileNetV2 
     & 70.97 & 59.24  \\
Ours--MobileNetV2
    & \textbf{72.88} & \textbf{62.89}  \\
\hline
Strided Conv.--MobileNetV2-0.35x
    & 56.64 & 49.20  \\
LIP--MobileNetV2-0.35x
    & 58.24 & 49.95  \\
GaussianPool--MobileNetV2-0.35x
     & 59.26 & 49.91  \\
RNNPool--MobileNetV2-0.35x 
  & 57.80 & 49.10  \\
Ours--MobileNetV2-0.35x
    & \textbf{60.92} & \textbf{51.16}  \\
\hline
\hline
\end{tabular}
\end{threeparttable}
}
\vspace{-10mm}
\end{center}
\end{table}

\subsection{Qualitative Results \& Visualization}
\begin{figure*}[htbp]
\centerline{\includegraphics[scale=0.62]{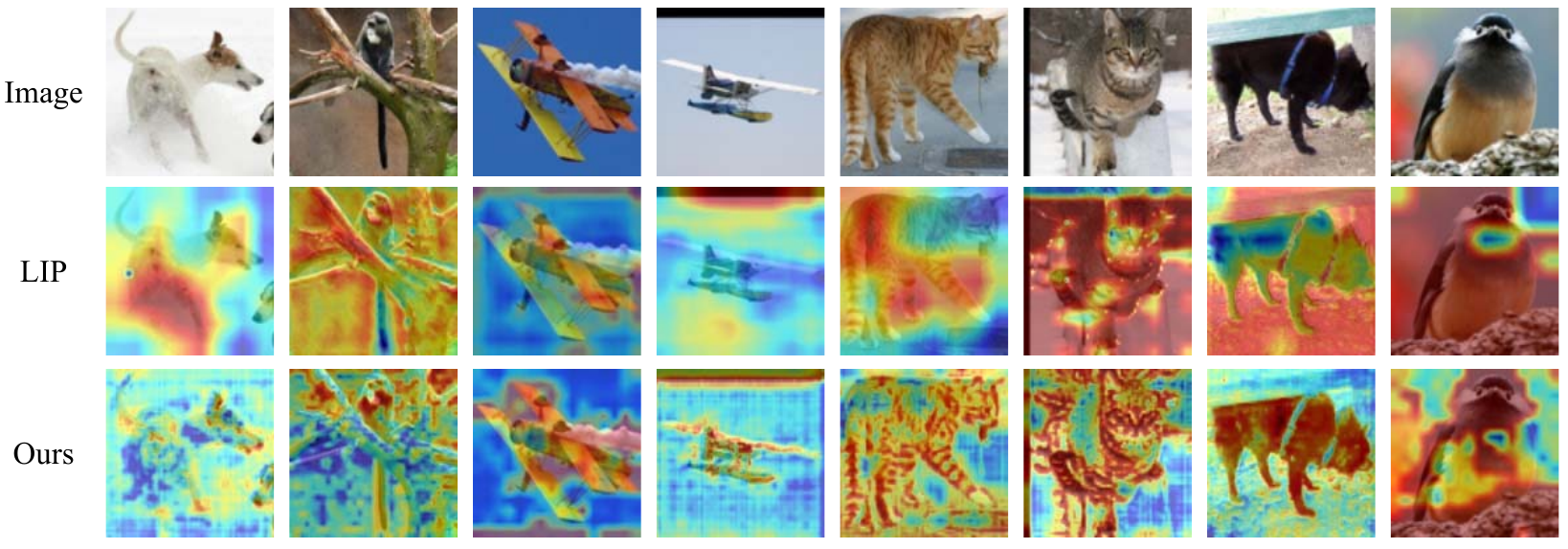}}
\caption{
 Visualization results for local importance based pooling and the proposed non-local self-attentive pooling.
 The images are from the STL10 dataset and the heatmaps in each technique highlight the regions of interest, i.e., the regions with high heatmap value will be regarded as effective information and retained while down-sampling.
 }
\label{fig:visualization}
\end{figure*}

 To intuitively illustrate the superiority of the proposed method, we visualize the heatmap corresponding to different attention mechanisms onto the images from STL10 dataset, as shown in Fig. \ref{fig:visualization}.

 Specifically, the heatmap is calculated by GradCam~\cite{selvaraju2017grad}, that computes the gradient of the ground-truth class for each of the pooling layers.
 The heatmap value is directly proportional to the pooling weights at a particular location, which implies that the regions with high heatmap values contain effective features that are retained during down-sampling.
 Compared with LIP, the representative locality-based pooling method, our proposed method is more concerned about the details of an image and the long-range dependencies between different local regions.
 As shown in the first and the second columns, LIP focuses only on the main local regions with large receptive fields. 
 In contrast, our method focuses on the features from different local regions, such as the dog's mouth, ear, legs in the first column and the bird and branches in the second column.
 These non-local features are related and might be established long-rang dependencies for feature aggregation.
 As shown in the fifth and sixth columns, our pooling method mainly focuses on the texture of the cat's fur, which might be a discriminative feature for classification/detection, while LIP focuses on the general shape of a cat. 
 This kind of general information might fail to guide feature aggregation when required to compress and retain effective detailed information.

\begin{table}
\caption{Comparison of the total FLOPs count of the whole CNN backbone with different pooling methods on the STL10 dataset.}
\label{tab:compute_cost}
\begin{center}
\setlength{\tabcolsep}{0.6mm}{
\begin{tabular}{l|c|c|c|c}
\hline
\hline
Architecture & Ours (G) & LIP (G) & GP (G) & Sd. Conv. (G) \\
\hline
MbNetV2 & 0.272 & 0.264  & 0.295 & 0.303 \\
\hline
MbNetV2-0.35x & 0.06 & 0.061 & 0.059 & 0.065 \\
\hline
ResNet18 & 1.82 & 1.93 & 1.77  & 2.07 \\
\hline
\hline
\end{tabular}}
\vspace{-8mm}
\end{center}
\end{table}



\subsection{Compute \& Memory Efficiency}\label{subsec:comp_mem}

Assuming the same input and output dimensions for down-sampling, and denoting the FLOPs count of our self-attentive pooling and the SOTA LIP layer as $F_{SA}$ and $F_{LIP}$ respectively, $F_{SA}\approx\frac{3}{n^2}F_{LIP}$. Hence, adopting a patch size $n>1$ makes our pooling costs cheaper than that of LIP. In particular, we use higher patch sizes (ranging from 2 to 8) for the initial pooling layers and a patch size of $1$ for the later layers (see Table \ref{tab:hyper parameter settings}). This still keeps our total FLOPs count of the entire model  lower than LIP, as shown in Table \ref{tab:compute_cost}, because the FLOPs count of both the pooling methods is significantly higher in the initial layers compared to the later layers due to the large size of the activation maps. Note that, in most standard backbones, the channel dimension only increases by a factor of $2$, when each of the spatial dimension reduces by a factor of $2$, which implies that the total size of the activation map progressively reduces as we go deeper into the network. Our method also consumes $11.66\%$ lower FLOPs, on average, compared to strided convolution based pooling, as shown in Table \ref{tab:compute_cost}.

The memory consumption of the whole CNN network is similar for both self-attentive pooling and LIP with identical backbone configurations and identical down-sampling in the pooling layers. Though our self-attentive pooling consists of significantly more trainable parameters for the query, key, and value computation compared to local trainable pooling layers, they are fixed during inference, and can be saved off-line in the on-chip memory. Also, the memory consumed by these parameters is still significantly lower compared to that by the initial activation maps, and hence, it does not  significantly increase the memory overhead. Note that the reduction of $s1$ by a factor of 2 approximately halves the total memory consumption, enabling the CNN deployment in devices with tighter memory budgets. As illustrated in Tables \ref{tab: comparison on stl10}, \ref{tab: comparison on vww}, and \ref{tab: comparison on imagenet}, the accuracy gain of our proposed pooling method compared to the SOTA grows as we increase $s1$. A similar trend is also observed as we go from MobileNetV2 to MobileNetV2-0.35x to reduce the memory consumption. For example, the accuracy gain further increases from $0.25\%$ to $4.97\%$ when evaluated on STL10 which implies the non-local self-attention map can extract more discriminative features from a memory-constrained model. \rev{For ImageNet, with the aggressive down-sampling of the activation maps in the initial layers (providing up to 22$\times$ reduction in memory consumption where 11$\times$ is due to MobileNetV2-0.35x and 2$\times$ is due to aggressive striding), the test accuracy gap with the SOTA techniques at iso-memory increases from $1.2\%$ on average to $1.43\%$.} All these motivate the applicability of our approach in resource-constrained devices. Channel pruning can further reduce the memory consumption of our models without too much reduction in the test accuracy. 
We consider 2$\times$ channel pruning in the 1$^{st}$ stage of all the backbone networks, as illustrated in Table~\ref{tab: comparison on stl10} and \ref{tab: comparison on vww}. As we can see, addition of channel pruning with $s1{=}1$ can retain (or sometimes even outperform) the accuracy obtained by our proposed pooling technique. However, channel pruning does not improve the accuracy obtained for more aggressive down-sampling with our pooling technique ($s1{=}2,4$). Hence, the nominal down-sampling schedule ($s1{=}1$) with channel pruning is the most suitable configuration to reduce the memory footprint.

\subsection{Ablation Study}\label{subsec:ablate}
 We conduct ablation studies of our proposed pooling method when evaluated with ResNet18 backbone on the STL10 dataset. 
 Our results are shown in Table~\ref{tab: ablation study}.
 Note that bn1 and bn2 denote the BN layers in the patch embedding and multi-head self-attention modules respectively, and pe denotes the positional encoding layer.
 \rev{SelfAttn directly uses the muti-head self-attention module before each strided convolution layer without spatial-channel restoration and weighted pooling.}
 Removing either of the BN layer results in a slight drop in test accuracy.
 \rev{We hypothesize the batch norm (BN) layers normalize the input data distribution, which helps the non-linear activation extract better features and help speed up convergence. Note that this argument is valid for BN layers in CNNs, not particular to self-attentive pooling.}
 Our pooling method without exponential function degenerates significantly.
 This might be because each value in the attention map after the sigmoid function is limited in $0\sim1$, without amplifying the response of effective features.
 Removal of the positional encoding also slightly reduces the accuracy which illustrates the importance of positional information.
 \rev{We hypothesize the position encoding layer merges the positional information into patch tokens, thereby compensating the broken spatial relationship between different tokens.}
 Also, our pooling method without sigmoid yields only statistical test accuracy. This is because, without the sigmoid rectification, the output of the spatial-channel restoration module goes to infinity after the amplification by exponential function, resulting in gradient explosion. \rev{Compared to using only self-attention module (instead of our proposed pooling technique) before the strided convolution, our proposed method is more effective. As illustrated in Table \ref{tab: ablation study},  our accuracy increase is due to the proposed methods, not only the self-attention mechanism.\footnote{We do not find the self-attention module to be effective probably because we do not pre-train it on large datasets, such as JFT-300M \cite{dosovitskiy2020image}.}
  }

\begin{table}[t]
\caption{Ablation Study of our Proposed Pooling Technique.}
\label{tab: ablation study}
\begin{center}
\setlength{\tabcolsep}{0.5mm}{
\begin{threeparttable}
\begin{tabular}{l|c|c}
\hline
\hline
Metrics & \multicolumn{2}{c}{Top 1 Acc. (\%)} \\
\hline

\diagbox[]{Methods}{$1^{st}$ Pool Stride} 
  & $s1=1$ & $s1=2$\\


\hline
w$\setminus$o(bn1)--ResNet18 (Outer Stage) & 80.34 & 78.45 \\
w$\setminus$o(bn2)--ResNet18 (Outer Stage) & 82.01 & 80.36 \\
w$\setminus$o(exp)--ResNet18 (Outer Stage) & 81.95 & 80.00 \\
w$\setminus$o(pe)--ResNet18 (Outer Stage) & 82.01 & 79.73 \\
w$\setminus$o(sigmoid)--ResNet18 (Outer Stage) & 10.00 & 10.00 \\
SelfAttn-MobileNetV2 (Inner Stage) & 13.44 & - \\
SelfAttn-MobileNetV2 (Outer Stage) & 13.23 & - \\
SelfAttn-ResNet18 (Inner Stage) & 26.17 & - \\
SelfAttn-ResNet18 (Outer Stage) & 58.71 & - \\
\hline
Ours--ResNet18 (Outer Stage)
    & \textbf{82.25} & \textbf{81.11} \\
Ours--ResNet18 (Inner Stage) & 81.45 & 79.17 \\

Ours--MobileNetV2 (Outer Stage) & 79.45 & 68.81 \\
Ours--MobileNetV2 (Inner Stage) & \textbf{81.75} & \textbf{75.39} \\

\hline
\hline
\end{tabular}
\end{threeparttable}
}
\vspace{-4mm}
\end{center}
\end{table}


\section{Conclusion \& Societal Implications}

In this paper, we propose self-attentive pooling which aggregates non-local features from the activation maps, thereby enabling the extraction of more complex relationships between the different features, compared to existing local pooling layers. Our approach outperforms the existing pooling approaches with popular memory-efficient CNN backbones on several object recognition and detection benchmarks. Hence, we hope that our approach can enable the deployment of accurate CNN models on various resource-constrained platforms such as smart home assistants and wearable sensors. While our goal is to enable socially responsible use-cases, our work can also unlock several cheap and real-time vision use-cases that might be susceptible to adversarial attacks and racial biases. 
Preventing the application of this technology from abusive usages is an important area of future work.

\section{Acknowledgements}

We would like to acknowledge the DARPA HR$00112190120$ award and the NSF CCF-$1763747$ award for supporting this work. The views and conclusions contained herein are those of the authors and should not be interpreted as necessarily representing the official policies or endorsements of DARPA or NSF.

{\small
\bibliographystyle{ieee_fullname}
\bibliography{egbib}
}

\end{document}